\documentclass[10pt,twocolumn,letterpaper]{article}

\usepackage{cvpr}              

\usepackage[misc]{ifsym}
\newcommand\blfootnote[1]{
  \begingroup
  \renewcommand\thefootnote{}\footnote{#1}
  \addtocounter{footnote}{-1}
  \endgroup
}

\usepackage[dvipsnames]{xcolor}

\definecolor{cvprblue}{rgb}{0.21,0.49,0.74}
\usepackage[pagebackref,breaklinks,letterpaper,colorlinks,citecolor=cvprblue]{hyperref}
\usepackage[capitalize]{cleveref}

\def\paperID{***} 
\def\confName{CVPR}
\def\confYear{2024}

\title{Drag-A-Video: Non-rigid Video Editing with Point-based Interaction}

\author{
Yao Teng\textsuperscript{1} \quad Enze Xie\textsuperscript{2~$\dagger$} \quad Yue Wu\textsuperscript{2}  \quad Haoyu Han\textsuperscript{3} \quad Zhenguo Li\textsuperscript{2} \quad Xihui Liu\textsuperscript{1~$\dagger$} \\
\textsuperscript{1}The University of Hong Kong  \quad
\textsuperscript{2}Huawei Noah’s Ark Lab  \quad \textsuperscript{3}Tsinghua University \\
}

\begin{document}

\twocolumn[{
\renewcommand\twocolumn[1][]{#1}
\maketitle

\begin{center} 
    \includegraphics[width=0.999\linewidth]{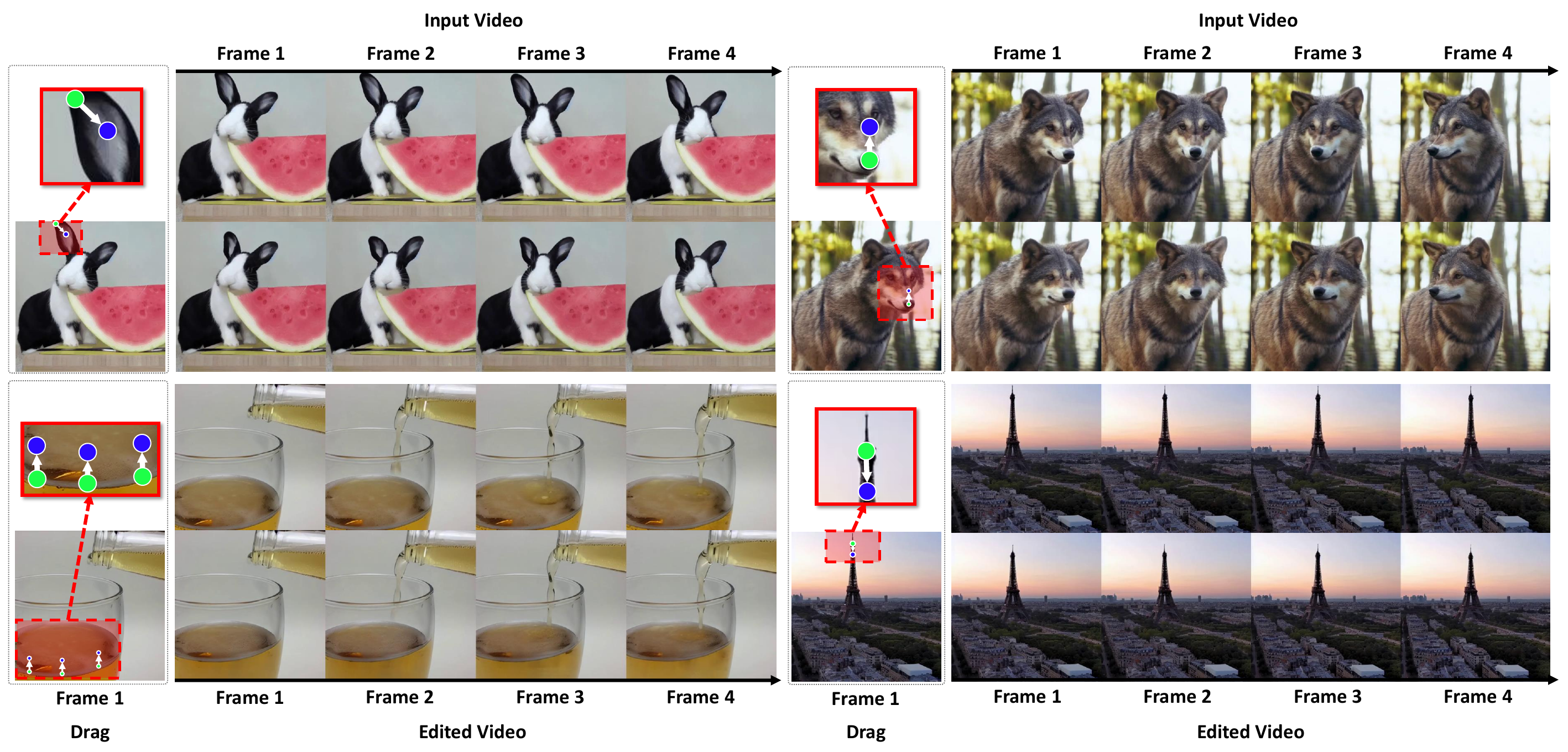}
    \captionsetup{type=figure}
    \vspace{-2em}
    \caption{
    \textbf{Drag-A-Video enables the point-based manipulation on videos}. Drag with multiple points can be realized on the video data. The green and blue points denote the handle and target points, respectively. The red regions are for better viewing.
    }
    \label{fig:teaser}
    \vspace{.5em}
\end{center}
}]

\begin{abstract}
Video editing is a challenging task that requires manipulating videos on both the spatial and temporal dimensions. Existing methods for video editing mainly focus on changing the appearance or style of the objects in the video, while keeping their structures unchanged. However, there is no existing method that allows users to interactively ``drag'' any points of instances on the first frame to precisely reach the target points with other frames consistently deformed. In this paper, we propose a new diffusion-based method for interactive point-based video manipulation, called Drag-A-Video. Our method allows users to click pairs of handle points and target points as well as masks on the first frame of an input video. Then, our method transforms the inputs into point sets and propagates these sets across frames. To precisely modify the contents of the video, we employ a new video-level motion supervision to update the features of the video and introduce the latent offsets to achieve this update at multiple denoising timesteps. We propose a temporal-consistent point tracking module to coordinate the movement of the points in the handle point sets. We demonstrate the effectiveness and flexibility of our method on various videos. The website of our work is available here: {\url{https://drag-a-video.github.io/}}.
\end{abstract}
\blfootnote{$\dagger$ Corresponding authors.}
\section{Introduction}
\label{sec:intro}
The advancement of diffusion-based generative models has showcased exceptional capabilities in producing diverse and photorealistic images~\cite{stable_diffusion,imagen} and even videos~\cite{make-a-video,zhou2022magicvideo,ho2022video, hong2022cogvideo,wang2023videofactory,text2videozero, imagenvideo} conditioned on text. Recently, the editing and variation of existing videos with diffusion-based generative models have gained significant attention. Prior approaches primarily centered around text-driven video editing~\cite{fatezero,tuneavideo,stablevideo,tokenflow,FLATTEN}. However, these methods were limited in their controllability, relying solely on textual descriptions for conveying desired edits. 
Furthermore, they typically facilitated global changes such as style transfer, lacking precise and fine-grained control. Consider, for instance, social media users who might seek to intricately adjust the shape, pose, or structure of a specific object or the structures or layouts within a video. Such a detailed control is challenging to achieve using only text descriptions.
Recent work DragGAN~\cite{draggan} enables users to apply drag-based image manipulation with the input control points interactively.
The dragging process is carried out by alternately optimizing the latent code of the image and updating the control points.
Drawing inspiration from DragGAN~\cite{draggan}, we explore \textit{\textbf{interactive point-based video manipulation}}, where the goal is to \textit{allows users to drag points only on the first frame and the other frames of the whole video clip will be deformed consistently by our algorithm.} 

There are three main challenges in designing such a point-based video editing algorithm.
\textit{First}, it is difficult to propagate the input control points across frames considering the inherent motion in videos. 
\textit{Second}, preserving the spatial coherency within each frame is challenging. 
\textit{Third}, manipulating videos by dragging entails the temporal consistency challenge, which necessitates model designs that incorporate temporal consistency for latent optimization and control point updating.
In conclusion, point-based video editing is challenging and can not be solved by simply extending existing video editing or point-based image editing techniques.

To mitigate those problems, we introduce \textit{\textbf{Drag-A-Video}}, the first point-based, interactive, non-rigid video editing system. 
Our method allows for precise control over object structure and the simulation of non-rigid motion dynamics, yielding highly consistent results.
Specifically, given a video input, we allow users to click pairs of handle points and target points, indicating that the handle points should move towards the target point locations during editing. An optional mask can be provided by the user to constrain that only the masked region should be edited.
Our pipeline comprises three components: point set propagation, latent optimization with motion supervision, and temporal-consistent point tracking. Given an input video with the user-defined handle points, target points, and mask on the first frame, we first propagate the point sets and mask to other frames to facilitate frame-wise editing. Instead of propagating only the user-defined points, we propagate a larger point set to ensure robust dragging in all frames. We then alternately optimize the diffusion latents with video-level motion supervision and update handle points based on the optimized latents with temporal-consistent point tracking. Different from previous approaches, which only optimize the diffusion latents at a fixed timestep, our approach can optimize diffusion latents of multiple timesteps by introducing learnable offset maps for the latents. Both video-level motion supervision and temporal-consistent point tracking introduce temporal consistency constraints to enable coherent video editing.

In summary, we present Drag-A-Video, the first point-based interactive non-rigid video editing framework. Experiments demonstrate that our design conducts high-quality and temporal-consistent point-based video editing based on the drag control only at the first frame.
\begin{figure*}
    \centering
    \vspace{-1.5em}
    \includegraphics[width=0.95\linewidth]{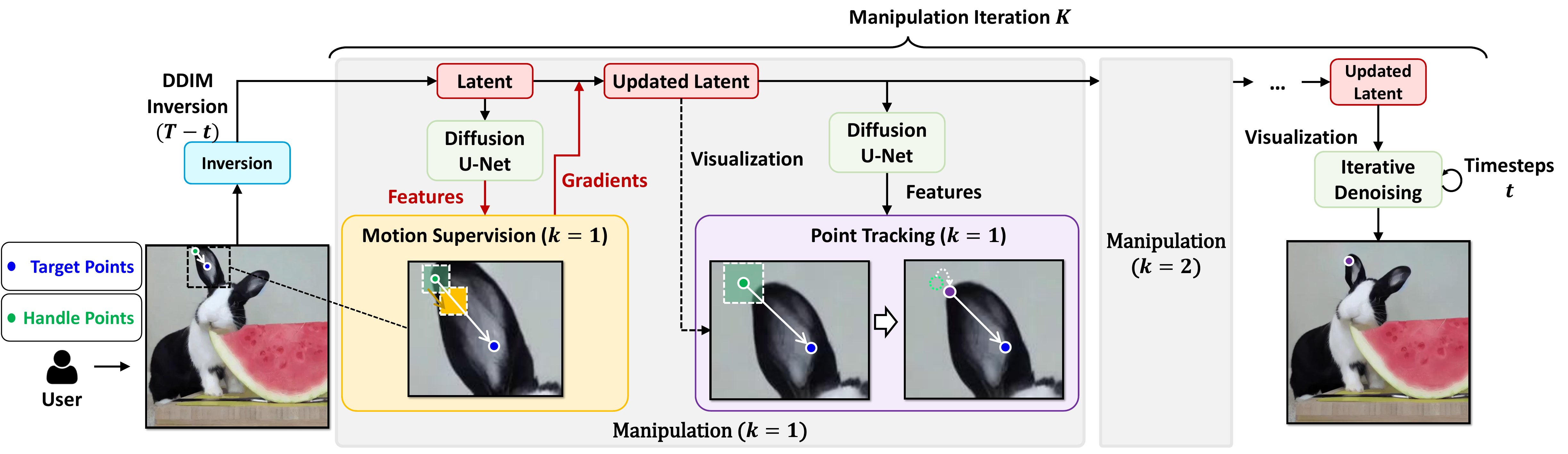}
    \vspace{-.5em}
    \caption{
    {The pipeline of the point-based image editing on the diffusion model}.
    This algorithm is an iterative optimization process composed of motion supervision and point tracking. The motion supervision provides gradients to update the latent. The point tracking updates the location of each handle point with a small step toward the target point.
    }
    \vspace{-.5em}
    \label{fig:dragdiff}
\end{figure*}

\section{Related Work}
\label{sec:relatedwork}
\textbf{Image and Video Generation}.
Image generation has been a long-standing challenge in computer vision.
Many models have been proposed over the years~\cite{gan,stylegan}.
More recently, diffusion models~\cite{ddpm,ddpm2,edm,cold_diffusion,unified_perspective_diffusion,ddim,beats_gan} have made a breakthrough in high-quality text-to-image generation~\cite{stable_diffusion,dalle,attribute_t2i,imagen}.
These models denoise a random Gaussian noise into a realistic image in an iterative manner.

\noindent
\textit{\textbf{Text-to-Image Diffusion Model}}. 
The progression of the text-to-image (T2I) generative models, including DALL·E~2~\citep{Dalle-2}, Imagen~\citep{imagen}, Stable Diffusion~\citep{stable_diffusion}, Midjourney~\citep{Midjourney} marks the onset of a groundbreaking phase in photorealistic image synthesis. Leveraging the potential of these powerful pretrained T2I models for image editing remains a challenging and unresolved problem.

\noindent
\textit{\textbf{Text-to-Video Diffusion Model}}.
Advancements have been achieved by applying diffusion models on text-to-video (T2V) synthesis. A series of studies have concentrated on enhancing the quality of generated videos~\cite{ho2022video, hong2022cogvideo,wang2023videofactory,make-a-video,text2videozero, imagenvideo}. Notably, Make-A-Video~\cite{make-a-video} transfers the internal knowledge of T2I models to video generation in an unsupervised training manner. Text2Video-Zero~\cite{text2videozero} also leveraged the power of the T2I model and proposed the latent code enrichment and new cross-attention attention to achieve motion dynamics and consistency. Imagen Video~\cite{imagenvideo} used a cascade of video diffusion models. Another stream of works focuses on video synthesis with additional control signals~\cite{videocomposer,chen2023controlavideo}.

\noindent
\textbf{Video Editing}.
Layered Neural Atlases~(LNA)~\cite{atlas} learned to map the frame pixels into layered 2D atlases to obtain unified foreground/background representation, enabling consistent video editing like texture mapping and style transfer.
Text2Live~\cite{text2live} combined the pre-trained LNA with CLIP~\cite{clip} for zero-shot text-driven video editing.
INVE~\cite{inve} introduces the hashing algorithm from InstantNGP~\cite{instantngp} to improve LNA.
There are also many video editing methods specifically designed for pre-trained diffusion models.
Tune-A-Video~\cite{tuneavideo} inflated operators in 2D text-to-image generative models to the spatio-temporal domain and then performed one-shot tuning on each video for text-driven video editing.
StableVideo~\cite{stablevideo} used LNA to achieve temporal consistency of diffusion video editing.
FateZero~\cite{fatezero} kept temporal consistency by using the attention maps stored in the inversion period. TokenFlow~\cite{tokenflow} utilized DIFT~\cite{dift} for feature propagation across the temporal dimension, while Pix2Video~\cite{pix2video} leveraged the gradients of the distance of the neighboring predicted clean frames to obtain consecutive results. 
Most of these previous video editing methods use text descriptions to drive the editing process, which is inadequate and can not provide precise and detailed control.

\noindent\textbf{Point-based Image and Video Editing}.
DragGAN~\cite{draggan} is the first interactive point-based image editing model that enables fine-grained spatial manipulation.
DragDiffusion~\cite{dragdiffusion} successfully introduced interactive point-based image editing from DragGAN to the diffusion models with one-shot fine-tuning. 
Concurrently, DragonDiffusion~\cite{dragondiffusion} is a tuning-free method that can perform not only point-based image editing but also more generalized image editing such as object moving and resizing.
FreeDrag~\cite{freedrag} designed an adaptive exponential moving average~(EMA) strategy to unify the motion supervision and the point tracking algorithm.
In this paper, we introduce point-based manipulation into the video domain and achieve consistent non-rigid video editing.
Unlike the previous methods, our Drag-A-Video is a non-rigid video editing algorithm that aims to edit the instance structures in videos rather than just modify the instance appearances. 

\begin{figure*}
    \centering
    \vspace{-1.5em}
    \includegraphics[width=0.99\linewidth]{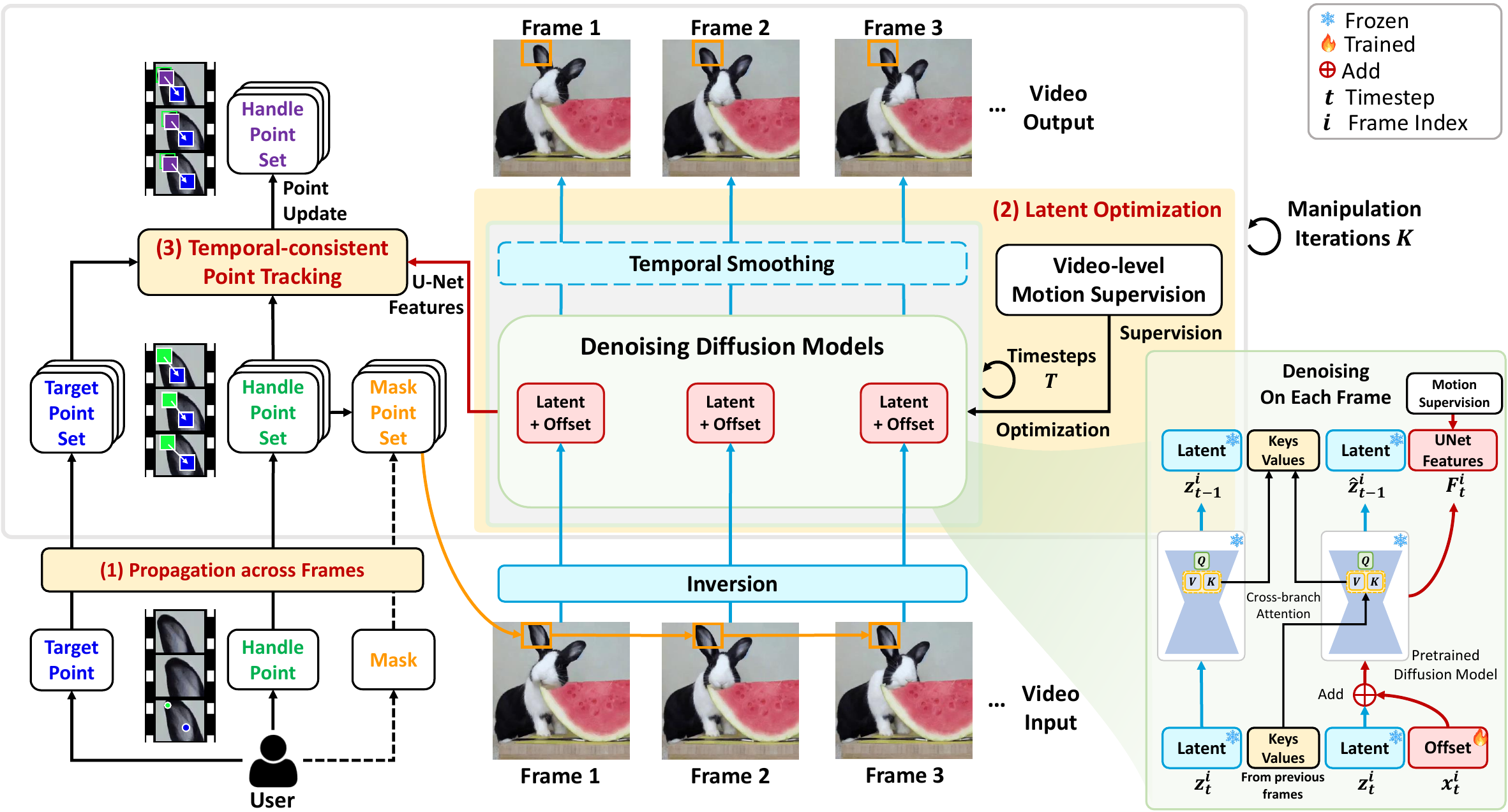}
    \vspace{-.5em}
    \caption{
    \textbf{Overview of Drag-A-Video}. The input is a video, as well as handle points, target points, and masks on the first frame.
    Our framework is an iterative optimization process. Each iteration contains a latent optimization and a temporal-consistent point tracking.
    In each latent optimization, the gradients provided by the video-level motion supervision are accumulated in the learnable latent offset modules, and are used to generate new images and features with the latent.
    In each point tracking, the handle points will be updated toward target points by a small step.
    After all the iterations are finished, we obtain the final edited videos.
    In this figure, the dashed lines indicate optional. 
    For simplicity, we omit the index $k$ on the variables.
    }
    \vspace{-.5em}
    \label{fig:dragavideo}
\end{figure*}

\section{Drag-A-Video}

As shown in \cref{fig:dragavideo}, given a video along with the user inputs, including handle points, target points, and masks on the first frame, our goal is to move the handle points to the target points on each video frame.
To achieve this, our model consists of three components. The point set propagation component propagates the handle points, target points, and masks defined by users from the first frame to other frames. We then alternately apply the latent optimization with video-level motion supervision and the temporal-consistent point tracking components. At each latent optimization step, we optimize the diffusion latent of each frame to drive the handle points to move toward the target points. The handle points are updated at each point tracking step based on the optimized diffusion latent.

In this section, we review point-based image editing in~\cref{subsec:pre}, propose the point set propagation algorithm in Sec.~\ref{sec:input}, present the latent optimization with motion supervision module in Sec.~\ref{sec:motion_supervision}, and introduce the temporal-consistent point tracking method in Sec.~\ref{sec:tracker}.

\subsection{Preliminaries: Point-based Image Editing}
\label{subsec:pre}
Pioneered by DragGAN~\cite{draggan}, point-based image editing is a brand-new technique in image editing. In contrast to other editing tasks, such as appearance replacement and style transfer, point-based image editing aims to deform the object structures or image layouts with precise and interactive control over the pixel locations.
During the editing process, the user clicks pairs of handle points and target points on an image, and then the model drives the semantic positions of the handle points to reach the corresponding target points while keeping the fidelity of the image.
Currently, most of the methods~\cite{dragdiffusion,draggan,freedrag,dragondiffusion} for point-based manipulation with diffusion models, as shown in~\cref{fig:dragdiff}, adopt an optimization-based approach consisting of two alternate steps: \textit{motion supervision and point tracking}.

\noindent\textbf{Motion supervision.}
In each motion supervision step, the latent codes of given images obtained by DDIM inversion~\cite{ddim} are optimized with the motion supervision that enforces the handle points to move towards the target points by feature distance loss. 
Additionally, a regularization loss is applied to force the unmasked region to be unchanged.

\noindent\textbf{Point tracking.}
In each point tracking step, the positions of the handle points are updated to track the corresponding points. The intermediate handle points are selected to minimize their feature distance from the original handle point in the input image.

With the alternate latent optimization and point update, the handle points iteratively move towards the target points. After $K$ optimization iterations, we denoise the optimized latents to generate the edited image.

\subsection{Point Set Propagation}\label{sec:input}
Since users only provide the handle points, target points, and mask for the first frame, 
we first propagate the points and mask to other frames to facilitate editing of all frames. The propagation is non-trivial because it requires robust and accurate tracking of the points and mask across frames.

\begin{figure}[t]
\centering
\begin{subfigure}{0.99\linewidth}
\centering
\includegraphics[width=0.9\linewidth]{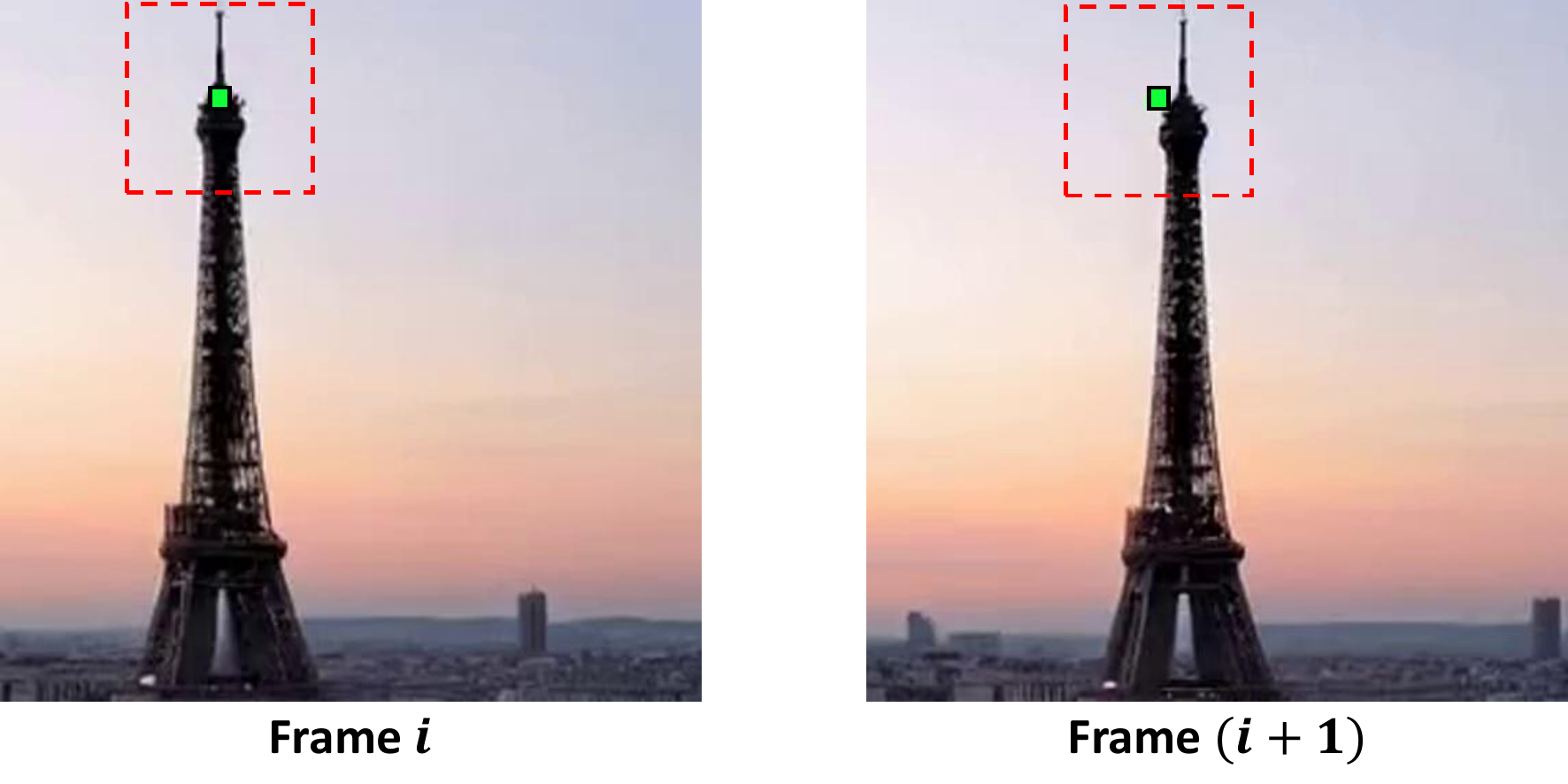}
\caption{The propagated handle point in the $(i+1)$-th frame may deviate from its expected location.
This figure shows a video shot around the Eiffel Tower where the handle point in the $i$-th frame is on the spire, but in the $(i+1)$-th frame, this point is incorrectly propagated to the background area around the spire instead of on the spire,
possibly affecting the dragging process.}
\label{fig:single_point}
\end{subfigure}
\\
\vspace{.5em}
\begin{subfigure}{0.99\linewidth}
\centering
\includegraphics[width=0.9\linewidth]{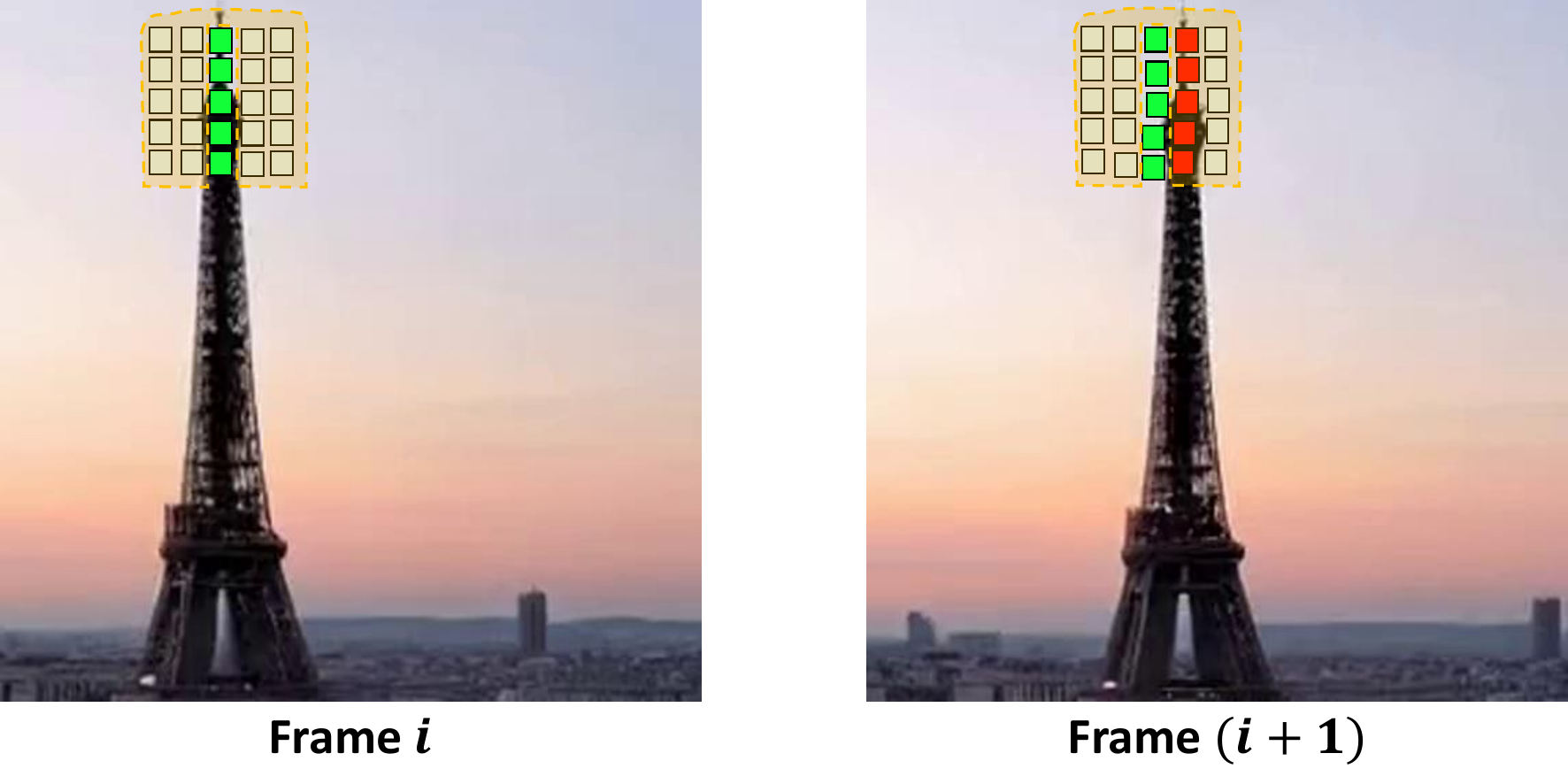}
\caption{Propagating a set of handle points instead of a single handle point improves robustness for dragging. Although the green points in the $i$-th frame are not accurately propagated to the tower's spire, the tower's spire in the $(i+1)$-th frame is covered by the red points, providing stable and correct dragging supervision.
}
\label{fig:multi_point}
\end{subfigure}
\caption{
An example of the point mapping.
(a) Traditionally, a single handle point is directly mapped from the first frame to another.
(b) We first extend the single handle point into a set and then perform mapping. 
} 
\vspace{-.5em}
\label{fig:point_num}
\end{figure}

\noindent\textbf{Robust handle point set propagation.}
To increase the robustness of our approach, instead of only propagating the handle points, we expand each handle point to a handle point set in a square patch centered at the handle point and propagate the whole set of points to other frames.
Since different points may have different motions in a video, the square patch on the first frame should be deformable when propagating to other frames. 
Given a set of points within the deformable patch in the $i$-th frame denoted as $\mathbf{P}^i$, we describe how we propagate the points to the $(i+1)$-th frame. We first leverage the Segment Anything Model (SAM)~\cite{sam} to distinguish the foreground points $\mathbf{P}^i_f$ from the background points $\mathbf{P}^i_b$. For the foreground points, we apply an existing algorithm DIFT~\cite{dift} to find the corresponding points on the next frame $\mathbf{P}^{i+1}_f$.
To increase the robustness, the background points are required to have the potential to cover instances in other frames.
Thus, each background point $\mathbf{p}^i_b$ is mapped to the next frame with the motion of its nearest foreground point $\hat{\mathbf{p}}^i_f$, denoted by the following equation,
\begin{equation}
    \mathbf{p}^{i+1}_b=\mathbf{p}^{i}_b+\hat{\mathbf{p}}^{i+1}_f-\hat{\mathbf{p}}^i_f.
\end{equation}
It is worth noting that our handle point set propagation allows for robust dragging on each frame.
For example, the propagated handle point in the $(i+1)$-th frame may deviate from its expected location.
\cref{fig:single_point} shows a video shot around the Eiffel Tower where the handle point in the $i$-th frame is on the spire, but in the $(i+1)$-th frame, this point is incorrectly propagated to the background area around the spire instead of on the spire,
possibly affecting the dragging process.
However, as shown in~\cref{fig:multi_point}, in our design,
although the green points in the $i$-th frame are not accurately propagated to the spire of the tower, the spire of the tower in the $(i+1)$-th frame is covered by the red points, providing stable and correct dragging supervision.

\noindent\textbf{Target point set propagation.} As introduced before, each handle point is paired with a target point to indicate the start and end point of dragging. The target point set in the $i$-th frame is denoted as $\mathbf{Q}^i$. Given a handle point $\mathbf{p}^i$ and the corresponding target point $\mathbf{q}^i$ on the $i$-th frame, as well as the propagated handle point on the $(i+1)$-th frame, we derive the coordinates of the propagated target by applying the same motion as the corresponding handle point,
\begin{equation}
    \mathbf{q}^{i+1}=\mathbf{q}^i+\mathbf{p}^{i+1}-\mathbf{p}^i.
\end{equation}

\begin{figure}[t]
\centering
\includegraphics[width=0.8\linewidth]{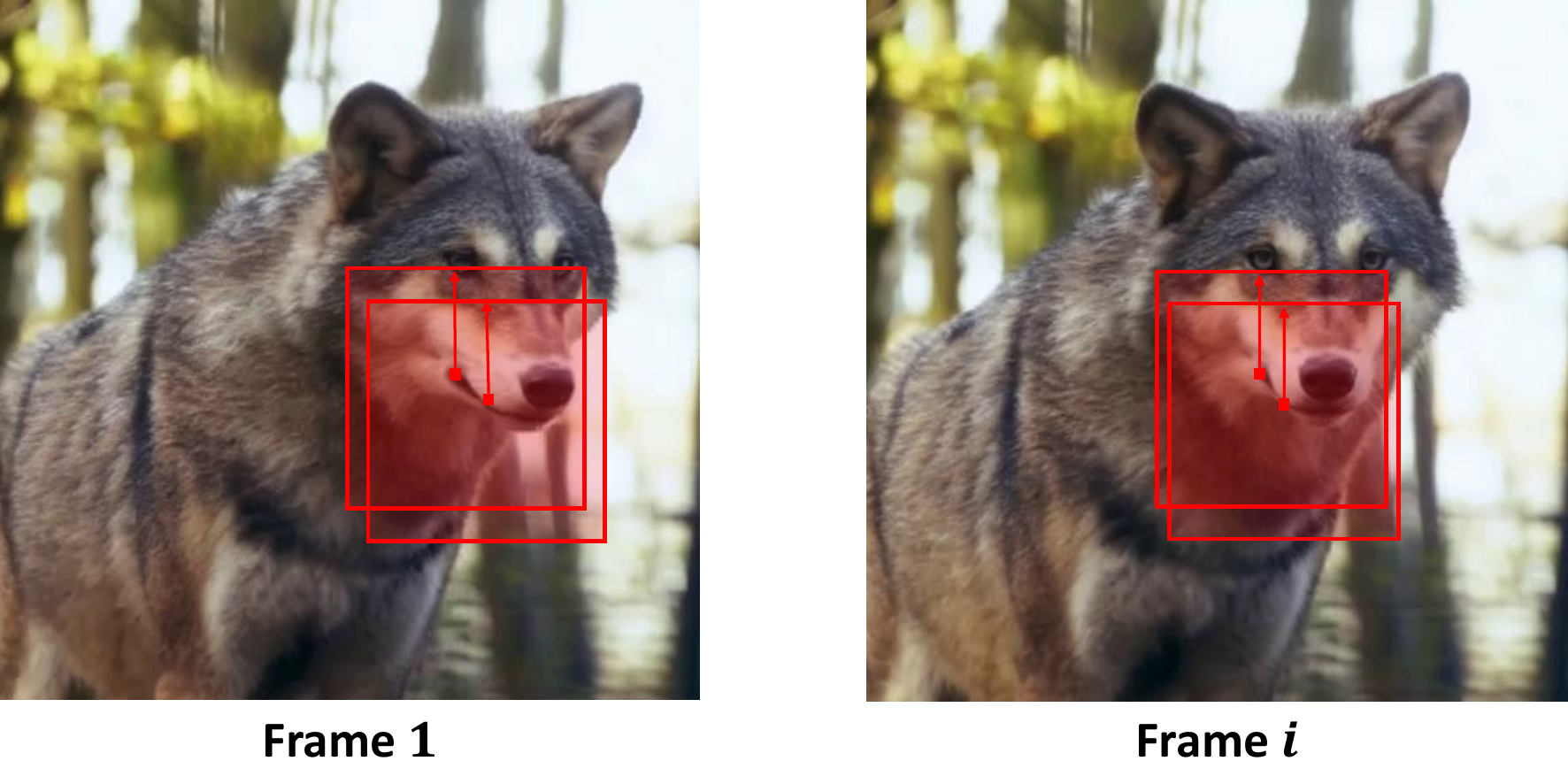}
\vspace{-.5em}
\caption{
An example of our mask region propagation.
Instead of inputting masks, the user inputs some mask points. These points are also extended into point sets, and then are mapped to other frames via DIFT~\cite{dift}.
The mask points are marked as red.
Each red square outlines a masked region, and the union of these regions forms the mask in~\cref{equ:lmask}.
}
\vspace{-1.2em}
\label{fig:mask}
\end{figure}

\noindent
\textbf{Mask Region Propagation}.
In point-based image and video editing, users can provide a mask to indicate the region that needs editing.
Since users cannot easily draw a precise mask boundary, and propagating each point in the mask across video frames is computationally heavy, we first simplify the mask by using a set of sparse points to represent the mask region. Specifically, we ask users to click on the first frame to define several ``mask points''. We propagate the mask points from the first frame to other frames in the same way as the handle point set propagation, and the mask region in each frame is defined as the union of the small patches centered at the mask points, as shown in~\cref{fig:mask}.

\subsection{Latent Optimization with Motion Supervision}\label{sec:motion_supervision}

After propagating the handle points, target points, and mask to all frames, we apply motion supervision on each frame to update the diffusion latent of each frame.
Specifically, each frame is first converted into the diffusion latent $\mathbf{z}_t$ by DDIM inversion~\cite{ddim}. 
Then, the diffusion latent is optimized with the motion supervision loss, driving the handle point towards the target point.
In this subsection, we introduce the video-level motion supervision to smoothly drag all frames consistently in videos and propose to optimize the latent across multiple diffusion timesteps to improve the controllability of dragging.

\noindent
\textbf{Video-level Motion Supervision.} 
In order to apply motion supervision to optimize diffusion latents of all video frames with temporal consistency, we propose to fuse the features of the target points across all frames for video-level motion supervision. 
Specifically, we denote a handle point at the $i$-th frame in the $k$-th optimization iteration as $\mathbf{p}_k^i$ and denote the corresponding target point as $\mathbf{q}^i$.\footnote{The motion supervision is applied on all points in the handle point set and target point set, but we only take one pair of handle point and target point for simplicity of notations.} Moreover, we denote the feature vector of the handle point extracted by the denoising U-Net as $F(\mathbf{p}_k^i)$.\footnote{In practice, we leverage the moving average of the feature vector over $k$ optimization iterations following FreeDrag~\cite{freedrag}.} To improve the temporal consistency, we average the handle point features across all frames,
\begin{equation}
    \bar{F}_k(\mathbf{p}) = \frac{1}{N}\sum_{i=1}^N F(\mathbf{p}^i_k), 
\end{equation}
where $N$ is the total number of frames in the video.
We denote the propagated mask in the $i$-th frame as $\mathbf{M}^i$. The drag loss for the target point in the $i$-th frame and $k$-th optimization iteration is,
\begin{equation}
    \mathcal{L}_{drag,k}^{i} = \Vert F(\mathbf{p}_k^i + \mathbf{d}_k^i) - \text{sg}(\bar{F}_k(\mathbf{p})) \Vert_1 ,
\label{equ:ldrag}
\end{equation}
where $\text{sg}(\cdot)$ denotes stop gradient, and $\mathbf{d}_k^i = \frac{\mathbf{q}^i-\mathbf{p}_k^i}{\Vert \mathbf{q}^i-\mathbf{p}_k^i \Vert_2}$ is the normalized vector from the handle point to the target point. Following previous work~\cite{draggan,freedrag,dragdiffusion}, the drag loss is computed over a small patch around the handle point $\mathbf{p}_k^i$, we omit this detail in the equation for ease of understanding. This loss drives the handle point to move towards the target point by a small step at each optimization iteration. The averaged handle point features $\bar{F}_k(\mathbf{p})$ improve temporal consistency by providing more stable and consistent supervision across frames.
Additionally, the mask loss is adopted following previous approaches to ensure the unmasked region remains unchanged. We denote the input of the diffusion model at the $t$-th timestep, $i$-th frame and the $k$-th optimization iteration as $\mathbf{x}_{t,k}^i$. The mask loss is computed as follows,
\begin{equation}
    \mathcal{L}_{mask,k}^{i} = \Vert (\mathbf{x}^i_{t,k} - \mathbf{x}^i_{t,0}) \odot (1 - \mathbf{M}^i ) \Vert_1 ,
\label{equ:lmask}
\end{equation}
where $\mathbf{M}^i$ is the propagated binary mask indicating the region to edit. 
Finally, the overall video-level motion supervision loss is computed as the summation of the drag loss and the mask loss.

\noindent
\textbf{Multi-timestep latent optimization of diffusion models.}
\label{sec:embed}
We use the aforementioned motion supervision to optimize the diffusion latent. Previous point-based image editing approaches on diffusion models~\cite{dragdiffusion,freedrag} optimize the diffusion latent at a fixed diffusion timestep. Specifically, they adopt DDIM inversion~\cite{ddim} for a fixed timestep $T^\prime$ to obtain the diffusion latent $\mathbf{z}_{T^\prime}$, and then optimize $\mathbf{z}_{T^\prime}$ with motion supervison.
However, we observe a trade-off between supervision effectiveness and editability of the latent.
For diffusion latent with small timesteps, the latent is closer to the clean image, so the motion supervision is more accurate, but the latent is less editable.
In contrast, for diffusion latent with large timesteps, the latent is noisy and it is difficult to extract features for the motion supervision precisely, but the noisy latent is more flexible for editing compared with clean latent.
Previous works try to find a balanced point in between empirically.
However, we propose a novel approach to optimize the diffusion latents at multiple timesteps.
Instead of directly optimizing the diffusion latents, we define an offset for each diffusion latent $\mathbf{z}_t$, and optimize the latent offset $\mathbf{x}_t$ for each timestep simultaneously. 
This design enables us to optimize the diffusion latents across multiple timesteps and improves the quality and controllability of the edits.

\noindent\textbf{Other techniques to improve temporal consistency.}
We further use more techniques to enhance the temporal consistency. Following~\cite{videocontrolnet,controlvideo,text2videozero}, we transform the self-attention modules into the \textit{cross-branch attention} for the upsampler layers of the denoising U-Net.
Additionally, following~\cite{videocontrolnet,controlvideo}, an optional temporal smoothing module is appended to the denoising U-Net.

\subsection{Temporal-consistent Point Tracking}\label{sec:tracker}

After optimizing the diffusion latents, the handle point locations are updated according to the new latents.
In an effort to achieve temporal-consistent dragging over the video, we impose a shared offset vector to update the handle points across different frames. The handle point update in the $k$-th optimization iteration is,
\begin{align}
    \Delta p_k &= \mathop{\arg\min}_{ \Delta p \in [-l, l] } \sum_{ i=1 }^{N} \Vert F(\mathbf{p}_k^i + \Delta p \cdot \mathbf{d}_k^i) - F(\mathbf{p}^i_k) \Vert_1 , \\
    \mathbf{p}_{k+1}^i &= \mathbf{p}_k^i + \Delta p_k \cdot \mathbf{d}_k^i,
\label{equ:glob_energy_pointtracking}
\end{align} 
where $\mathbf{d}_k^i = \frac{\mathbf{q}^i-\mathbf{p}_k^i}{\Vert \mathbf{q}^i-\mathbf{p}_k^i \Vert_2}$ is the unit vector from the handle point to the target point, and $l$ is the hyper-parameter controlling the maximum update range. In each update iteration, the handle points across different frames in the video are updated with the same offset $\Delta p_k$ to ensure the temporal consistency of edits.
\section{Experiments}

\begin{table}[t]
    \centering
    \small
    \begin{tabular}{c|ccc}
    \toprule
    Method  & Quality $\downarrow$ & Temp $\downarrow$ & Point $\downarrow$ \\
    \midrule
      Baseline  & 1.65 & 1.63 & 1.62\\
      Ours   & \textbf{1.35} & \textbf{1.37} & \textbf{1.38} \\
    \bottomrule
    \end{tabular}
    \vspace{-.5em}
    \caption{User Study. Our method achieves the best performance across the three aspects. Quality: frame quality. Temp: temporal consistency. Point: movement of handle points.}
    \vspace{-.5em}
    \label{tab:user_study}
\end{table}

\subsection{Implementation Details}
Our framework is based on SD1.5~\cite{stable_diffusion}.
We perform point-based manipulation on every frame, so each frame is equipped with a learnable latent offset in each selected timestep.
The selected timestep set $\mathcal{T}$ is $\{ 42,41,35,30 \}$.
We use Adam~\cite{adam} as our optimizer without weight decay, and the learning rate is set to 0.01 for latent optimization.
We use MasaCtrl~\cite{masactrl} as the implementation of our cross-branch attention.
The scalar $l$ in~\cref{equ:glob_energy_pointtracking} is set to $3$.
The maximal number of iterations of the manipulation is 60.
Following~\cite{dragdiffusion}, we use LoRA~\cite{lora} to fine-tune the attention modules of the pre-trained diffusion model with the video inputs.

\subsection{Qualitative Results}

\noindent
\textbf{Datasets}.
We currently select videos from LOVEU-TGVE~\cite{loveu_edit}, DAVIS~\cite{davis} and WebVid-10M~\cite{WebVid-10M}, and then click pairs of handle points and target points as well as masks on the first frame of the videos. The videos are resized and cropped into $512 \times 512$. Considering the computational resources, we sample 1 frame at an interval of 3 for manipulation.

\begin{figure*}
    \centering
    \vspace{-.2em}\includegraphics[width=0.9\linewidth]{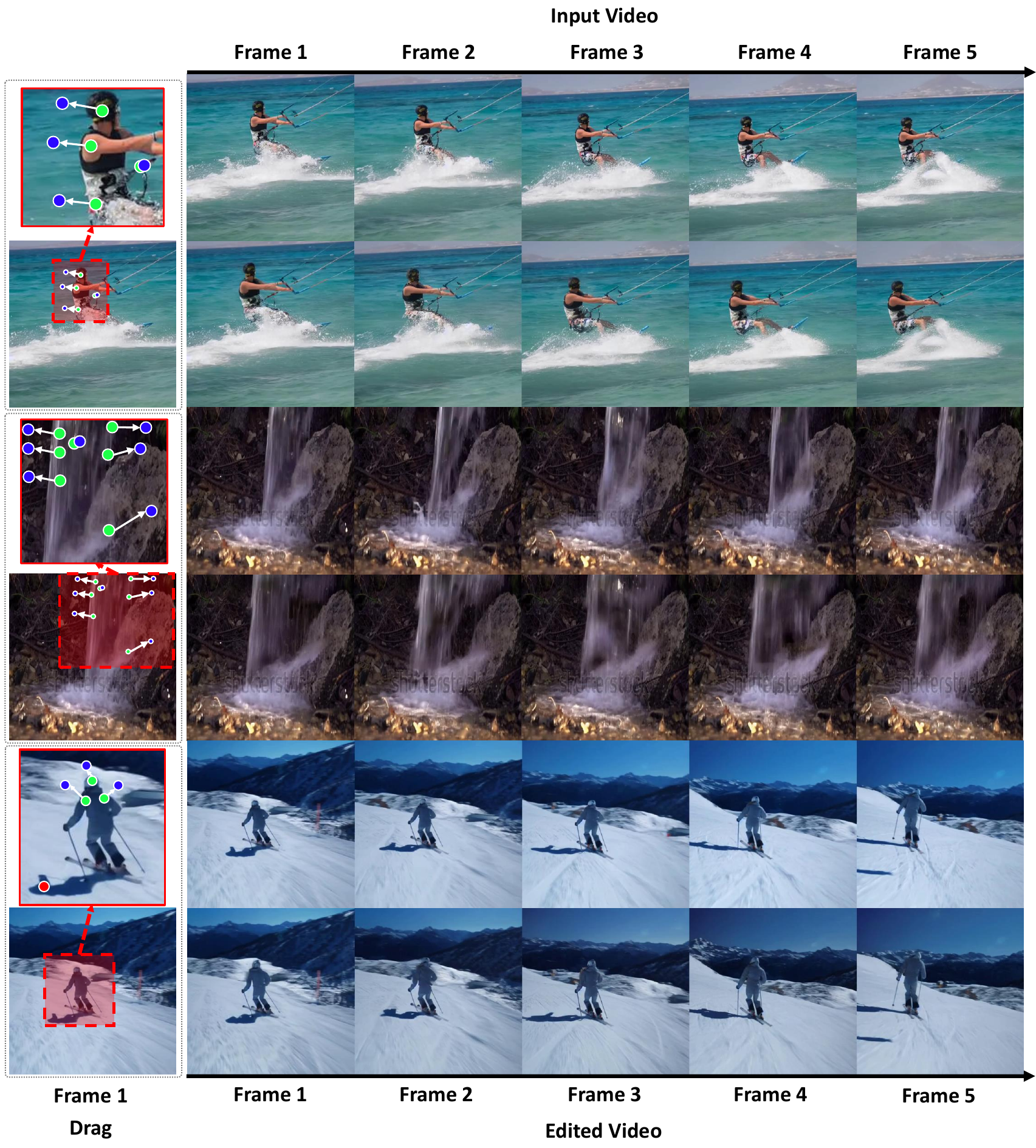}
    \vspace{-.7em}
    \caption{
    {\textbf{Results}}. The video samples were edited by our method. The green and blue points denote the handle and target points, respectively. The red point in the last case denotes the mask point. The video version of the results is in the appendix.
    }
    \vspace{-1em}
    \label{fig:sota}
\end{figure*}

\noindent
\textbf{Visualization}. Apart from~\cref{fig:teaser}, we present more edited videos in~\cref{fig:sota}. The qualitative results show that our model can drag the video contents from the input handle points toward target points, with the object structures changed. 
In addition to natural scenes where large objects take up much space (such as the waterfall in the second row), our framework is also practical on videos related to humans and small instances.

\subsection{User Study}

We perform a user study since human perception is more straightforward than automatic metrics. We asked 20 participants to evaluate the following three attributes of the edited videos: (a)~frame quality, which measures the clarity and realism of each frame; (b)~temporal consistency, which measures the smoothness and coherence of the video sequence, and (c)~the movement of handle points, which measures the accuracy of the user-expected motion of points.

We construct a baseline by removing the input propagation and the temporal consistency modules from our framework and using modules designed for individual images instead. We conduct a user study to compare our framework with this baseline, as shown in~\cref{tab:user_study}. We take the average ranking of the methods on the attributes as scores, and our method achieves the best performance across the three aspects.

\subsection{Ablation Studies}

\begin{figure}[t]
\vspace{-1.5em}
\centering
\includegraphics[width=0.99\linewidth]{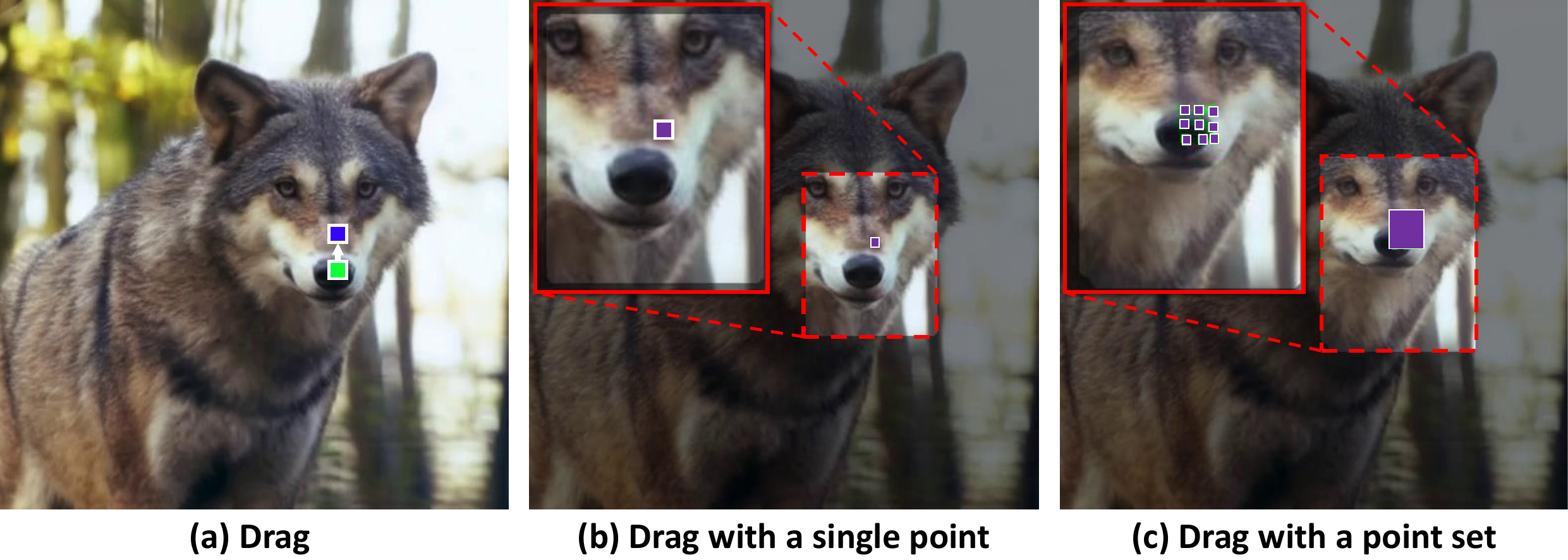}
\vspace{-.5em}
\caption{
Study of the point set. Using a single handle point (the squares marked in purple) as input can easily cause the lost tracking of points while using a point set can enhance the robustness of point-based manipulation.
} 
\vspace{-.5em}
\label{fig:abla_points}
\end{figure}

\begin{figure}[t]
\centering
\includegraphics[width=0.99\linewidth]{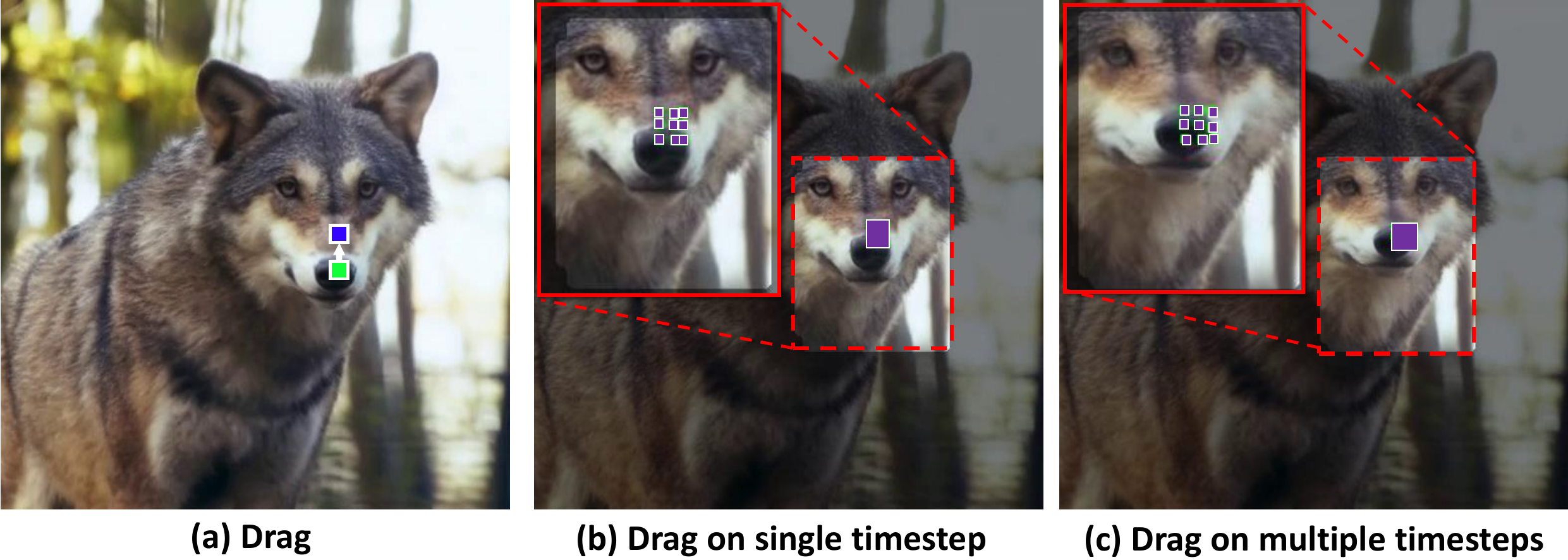}
\vspace{-.5em}
\caption{
Study of the manipulation on multiple timesteps. When the manipulation is performed on multiple timesteps, more points (the squares marked in purple) cover the nose of the wolf, so multiple timesteps lead to a better result.
} 
\vspace{-.5em}
\label{fig:abla_timestep}
\end{figure}

\begin{figure}[t]
\vspace{-1.5em}
\centering
\begin{subfigure}{0.99\linewidth}
\centering
\includegraphics[width=0.7\linewidth]{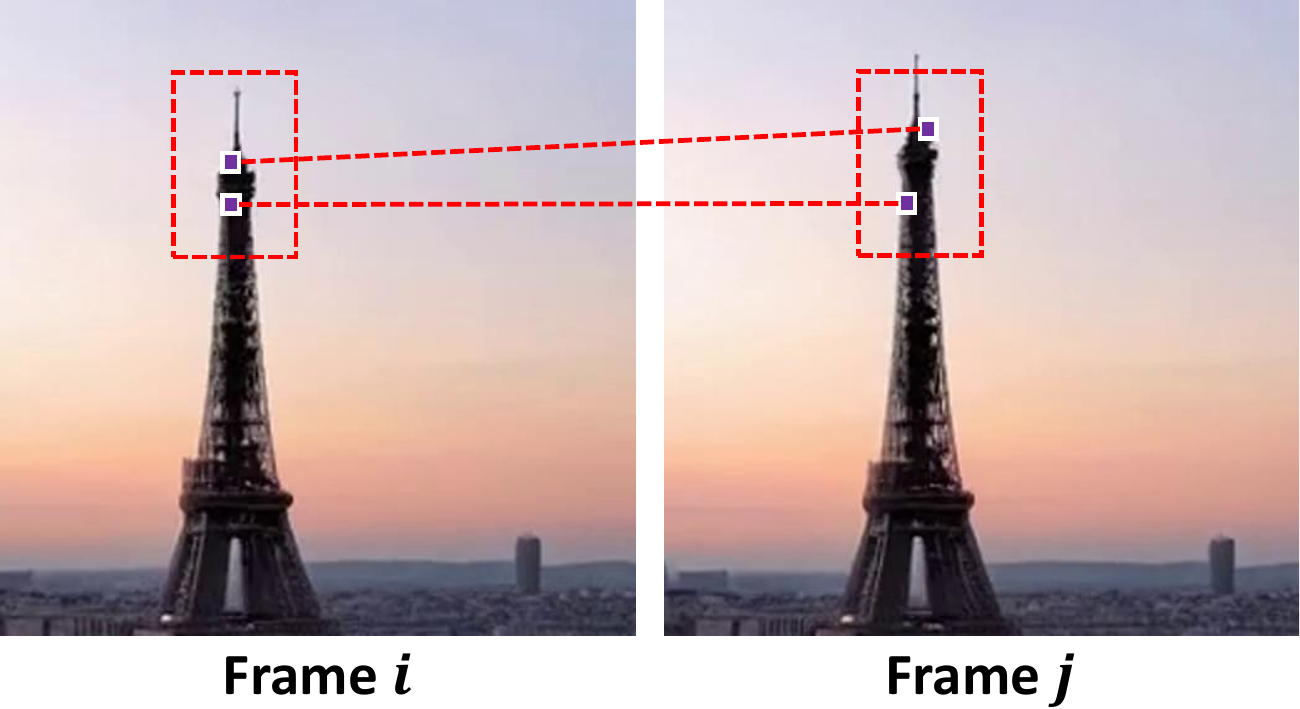}
\caption{
\textbf{Without our temporal-consistent modules}, the distance between two points on the first frame is different from the distance between two corresponding points on another frame (one of the dashed red lines is not horizontal).}
\label{fig:abla_tempcon_nocons}
\end{subfigure}
\\
\vspace{.5em}
\begin{subfigure}{0.99\linewidth}
\centering
\includegraphics[width=0.7\linewidth]{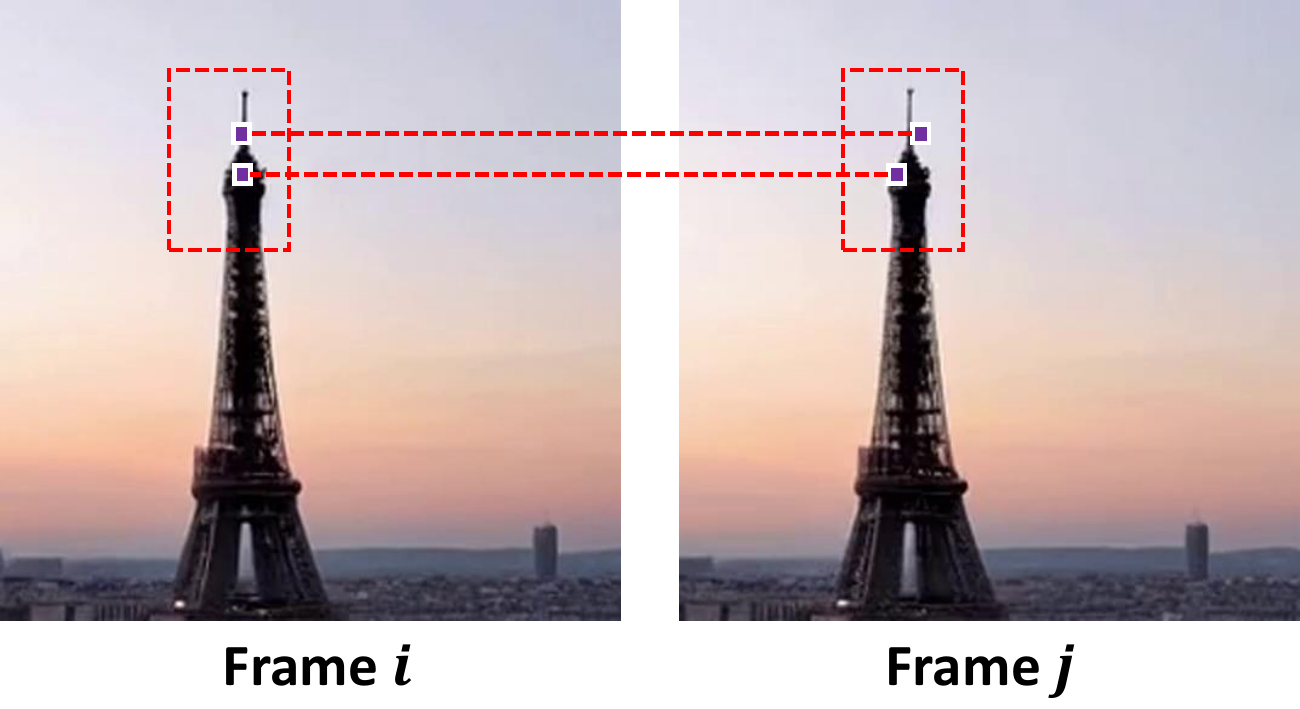}
\caption{ 
\textbf{With our temporal-consistent modules}, the distance between two points is consistent across frames (The dashed red lines are horizontal).
}
\label{fig:abla_tempcon_cons}
\end{subfigure}
\vspace{-.5em}
\caption{
Study on the temporal consistent modules. For visualization, we select two adjacent points located on the tower spire at the first frame with their corresponding points.
} 
\vspace{-.2em}
\label{fig:abla_temcons}
\end{figure}

\noindent
\textbf{Study on the point set}. 
We conduct ablation studies on the effectiveness of our proposed point set.
As shown in~\cref{fig:abla_points}, we compare the performance of our method with single or multiple handle points.
In \cref{fig:abla_points}~(b), if drag with only single initial \textcolor{ForestGreen}{handle point}, during editing, there is a high probability that the~\textcolor{DarkOrchid}{intermediate handle point} lose tracking the desired region. 
However, if we extend a single handle point to a point set~(\ie 3$\times$3 points), as shown in \cref{fig:abla_points}~(c), We can ensure that at least some of the points~(\ie 4/9 points) can cover the desired region.
In summary, we observed that extending a single point to a point set could significantly improve the robustness of our method, \ie, there is less risk of the failure of our point tracking.

\noindent
\textbf{Study on the multi-timestep manipulation}. Unlike the ordinary diffusion-based drag algorithm on a single timestep, our paradigm uses latent offsets to accumulate gradients on multiple timesteps. In~\cref{fig:abla_timestep}, we compare our method's performance applying single and multiple timesteps.
We click a handle point on the nose of the wolf. As shown in (b), when applying on a single timestep, the location of the~\textcolor{DarkOrchid}{intermediate handle point} shifts above the noise and can not accurately track the position of the nose. In (c), when the algorithm is performed on multiple timesteps, the center of the~\textcolor{DarkOrchid}{intermediate handle point} (the squares marked in purple) is much closer to the nose of the wolf.
Thus, the motion supervision on multiple timesteps leads to higher coherence between feature alternation and point movement.

\noindent
\textbf{Study on the temporal consistent designs}.
As shown in~\cref{sec:tracker}, our video-level motion supervision module and temporal-consistent point tracking module are tied together, which makes separate ablation studies to validate the effectiveness of these two modules not applicable. 
Consequently, we perform ablation studies on removing these two modules together, which will degrade our method to an image-based dragging method.
As shown in~\cref{fig:abla_temcons} (a), image-based editing on frame $j$ will lead the position of~\textcolor{DarkOrchid}{intermediate handle point} drift from the correct position, while in (b), with these two modules, our method can consistently track the correct positions of~\textcolor{DarkOrchid}{intermediate handle point} across frames. The results demonstrate our method can achieve strong temporal consistency.
More ablations are in the appendix.

\section{Conclusion}
In this paper, we propose a new challenge in video editing: point-based video manipulation. This challenge aims to modify the instance structures of videos given the point-based ``drag'' signals input by users on the first frame.
To handle the above challenge, we propose a baseline for point-based video manipulation, coined as Drag-A-Video.
In our framework, there are three components to ensure high-quality and temporal-consistent editing results: the user input propagation module, the multi-timestep latent optimization module with video-level motion supervision, and the temporal-consistent point-based tracking module.
Experiments demonstrate that our framework can achieve our goal, \ie, the contents of the first frames are dragged while other frames are consistently deformed. More discussions and limitations are provided in the appendix.
{
    \small
    \bibliographystyle{ieeenat_fullname}
    \bibliography{main}

\begin{thebibliography}{54}
\providecommand{\natexlab}[1]{#1}
\providecommand{\url}[1]{\texttt{#1}}
\expandafter\ifx\csname urlstyle\endcsname\relax
  \providecommand{\doi}[1]{doi: #1}\else
  \providecommand{\doi}{doi: \begingroup \urlstyle{rm}\Url}\fi

\bibitem[Bain et~al.(2021)Bain, Nagrani, Varol, and Zisserman]{WebVid-10M}
Max Bain, Arsha Nagrani, G{\"u}l Varol, and Andrew Zisserman.
\newblock Frozen in time: A joint video and image encoder for end-to-end retrieval.
\newblock In \emph{Proceedings of the IEEE/CVF International Conference on Computer Vision}, pages 1728--1738, 2021.

\bibitem[Baker et~al.(2011)Baker, Scharstein, Lewis, Roth, Black, and Szeliski]{smooth1}
Simon Baker, Daniel Scharstein, JP Lewis, Stefan Roth, Michael~J Black, and Richard Szeliski.
\newblock A database and evaluation methodology for optical flow.
\newblock \emph{International journal of computer vision}, 92:\penalty0 1--31, 2011.

\bibitem[Bansal et~al.(2022)Bansal, Borgnia, Chu, Li, Kazemi, Huang, Goldblum, Geiping, and Goldstein]{cold_diffusion}
Arpit Bansal, Eitan Borgnia, Hong{-}Min Chu, Jie~S. Li, Hamid Kazemi, Furong Huang, Micah Goldblum, Jonas Geiping, and Tom Goldstein.
\newblock Cold diffusion: Inverting arbitrary image transforms without noise.
\newblock \emph{CoRR}, abs/2208.09392, 2022.

\bibitem[Bao et~al.(2019)Bao, Lai, Zhang, Gao, and Yang]{smooth2}
Wenbo Bao, Wei-Sheng Lai, Xiaoyun Zhang, Zhiyong Gao, and Ming-Hsuan Yang.
\newblock Memc-net: Motion estimation and motion compensation driven neural network for video interpolation and enhancement.
\newblock \emph{IEEE transactions on pattern analysis and machine intelligence}, 43\penalty0 (3):\penalty0 933--948, 2019.

\bibitem[Bar-Tal et~al.(2022)Bar-Tal, Ofri-Amar, Fridman, Kasten, and Dekel]{text2live}
Omer Bar-Tal, Dolev Ofri-Amar, Rafail Fridman, Yoni Kasten, and Tali Dekel.
\newblock Text2live: Text-driven layered image and video editing.
\newblock In \emph{European conference on computer vision}, pages 707--723. Springer, 2022.

\bibitem[Cao et~al.(2023)Cao, Wang, Qi, Shan, Qie, and Zheng]{masactrl}
Mingdeng Cao, Xintao Wang, Zhongang Qi, Ying Shan, Xiaohu Qie, and Yinqiang Zheng.
\newblock Masactrl: Tuning-free mutual self-attention control for consistent image synthesis and editing.
\newblock \emph{arXiv preprint arXiv:2304.08465}, 2023.

\bibitem[Ceylan et~al.(2023)Ceylan, Huang, and Mitra]{pix2video}
Duygu Ceylan, Chun-Hao~P Huang, and Niloy~J Mitra.
\newblock Pix2video: Video editing using image diffusion.
\newblock In \emph{Proceedings of the IEEE/CVF International Conference on Computer Vision}, pages 23206--23217, 2023.

\bibitem[Chai et~al.(2023)Chai, Guo, Wang, and Lu]{stablevideo}
Wenhao Chai, Xun Guo, Gaoang Wang, and Yan Lu.
\newblock Stablevideo: Text-driven consistency-aware diffusion video editing.
\newblock In \emph{Proceedings of the IEEE/CVF International Conference on Computer Vision}, pages 23040--23050, 2023.

\bibitem[Chen et~al.(2023)Chen, Wu, Xie, Wu, Li, Xia, Xiao, and Lin]{chen2023controlavideo}
Weifeng Chen, Jie Wu, Pan Xie, Hefeng Wu, Jiashi Li, Xin Xia, Xuefeng Xiao, and Liang Lin.
\newblock Control-a-video: Controllable text-to-video generation with diffusion models, 2023.

\bibitem[Chu et~al.(2023)Chu, Lin, and Chen]{videocontrolnet}
Ernie Chu, Shuo-Yen Lin, and Jun-Cheng Chen.
\newblock Video controlnet: Towards temporally consistent synthetic-to-real video translation using conditional image diffusion models.
\newblock \emph{arXiv preprint arXiv:2305.19193}, 2023.

\bibitem[Cong et~al.(2023{\natexlab{a}})Cong, Min, Li, Rosenhahn, and Yang]{attribute_t2i}
Yuren Cong, Martin~Renqiang Min, Li~Erran Li, Bodo Rosenhahn, and Michael~Ying Yang.
\newblock Attribute-centric compositional text-to-image generation.
\newblock \emph{CoRR}, abs/2301.01413, 2023{\natexlab{a}}.

\bibitem[Cong et~al.(2023{\natexlab{b}})Cong, Xu, Simon, Chen, Ren, Xie, Perez-Rua, Rosenhahn, Xiang, and He]{FLATTEN}
Yuren Cong, Mengmeng Xu, Christian Simon, Shoufa Chen, Jiawei Ren, Yanping Xie, Juan-Manuel Perez-Rua, Bodo Rosenhahn, Tao Xiang, and Sen He.
\newblock Flatten: optical flow-guided attention for consistent text-to-video editing.
\newblock \emph{arXiv preprint arXiv:2310.05922}, 2023{\natexlab{b}}.

\bibitem[Dhariwal and Nichol(2021)]{beats_gan}
Prafulla Dhariwal and Alexander~Quinn Nichol.
\newblock Diffusion models beat gans on image synthesis.
\newblock In \emph{NeurIPS}, pages 8780--8794, 2021.

\bibitem[Ge et~al.(2022)Ge, Xu, Zhao, Itti, and Vineet]{dalle}
Yunhao Ge, Jiashu Xu, Brian~Nlong Zhao, Laurent Itti, and Vibhav Vineet.
\newblock {DALL-E} for detection: Language-driven context image synthesis for object detection.
\newblock \emph{CoRR}, abs/2206.09592, 2022.

\bibitem[Geyer et~al.(2023)Geyer, Bar-Tal, Bagon, and Dekel]{tokenflow}
Michal Geyer, Omer Bar-Tal, Shai Bagon, and Tali Dekel.
\newblock Tokenflow: Consistent diffusion features for consistent video editing.
\newblock \emph{arXiv preprint arXiv:2307.10373}, 2023.

\bibitem[Goodfellow et~al.(2014)Goodfellow, Pouget-Abadie, Mirza, Xu, Warde-Farley, Ozair, Courville, and Bengio]{gan}
Ian Goodfellow, Jean Pouget-Abadie, Mehdi Mirza, Bing Xu, David Warde-Farley, Sherjil Ozair, Aaron Courville, and Yoshua Bengio.
\newblock Generative adversarial nets.
\newblock \emph{Advances in neural information processing systems}, 27, 2014.

\bibitem[Ho et~al.(2020)Ho, Jain, and Abbeel]{ddpm}
Jonathan Ho, Ajay Jain, and Pieter Abbeel.
\newblock Denoising diffusion probabilistic models.
\newblock In \emph{NeurIPS}, 2020.

\bibitem[Ho et~al.(2022{\natexlab{a}})Ho, Chan, Saharia, Whang, Gao, Gritsenko, Kingma, Poole, Norouzi, Fleet, et~al.]{imagenvideo}
Jonathan Ho, William Chan, Chitwan Saharia, Jay Whang, Ruiqi Gao, Alexey Gritsenko, Diederik~P Kingma, Ben Poole, Mohammad Norouzi, David~J Fleet, et~al.
\newblock Imagen video: High definition video generation with diffusion models.
\newblock \emph{arXiv preprint arXiv:2210.02303}, 2022{\natexlab{a}}.

\bibitem[Ho et~al.(2022{\natexlab{b}})Ho, Salimans, Gritsenko, Chan, Norouzi, and Fleet]{ho2022video}
Jonathan Ho, Tim Salimans, Alexey Gritsenko, William Chan, Mohammad Norouzi, and David~J Fleet.
\newblock Video diffusion models.
\newblock \emph{arXiv:2204.03458}, 2022{\natexlab{b}}.

\bibitem[Hong et~al.(2022)Hong, Ding, Zheng, Liu, and Tang]{hong2022cogvideo}
Wenyi Hong, Ming Ding, Wendi Zheng, Xinghan Liu, and Jie Tang.
\newblock Cogvideo: Large-scale pretraining for text-to-video generation via transformers.
\newblock \emph{arXiv preprint arXiv:2205.15868}, 2022.

\bibitem[Hu et~al.(2021)Hu, Shen, Wallis, Allen-Zhu, Li, Wang, Wang, and Chen]{lora}
Edward~J Hu, Yelong Shen, Phillip Wallis, Zeyuan Allen-Zhu, Yuanzhi Li, Shean Wang, Lu Wang, and Weizhu Chen.
\newblock Lora: Low-rank adaptation of large language models.
\newblock \emph{arXiv preprint arXiv:2106.09685}, 2021.

\bibitem[Huang et~al.(2023)Huang, Sigal, Yi, Wang, and Lee]{inve}
Jiahui Huang, Leonid Sigal, Kwang~Moo Yi, Oliver Wang, and Joon-Young Lee.
\newblock Inve: Interactive neural video editing.
\newblock \emph{arXiv preprint arXiv:2307.07663}, 2023.

\bibitem[Jiang et~al.(2018)Jiang, Sun, Jampani, Yang, Learned-Miller, and Kautz]{smooth3}
Huaizu Jiang, Deqing Sun, Varun Jampani, Ming-Hsuan Yang, Erik Learned-Miller, and Jan Kautz.
\newblock Super slomo: High quality estimation of multiple intermediate frames for video interpolation.
\newblock In \emph{Proceedings of the IEEE conference on computer vision and pattern recognition}, pages 9000--9008, 2018.

\bibitem[Karras et~al.(2019)Karras, Laine, and Aila]{stylegan}
Tero Karras, Samuli Laine, and Timo Aila.
\newblock A style-based generator architecture for generative adversarial networks.
\newblock In \emph{Proceedings of the IEEE/CVF conference on computer vision and pattern recognition}, pages 4401--4410, 2019.

\bibitem[Karras et~al.(2022)Karras, Aittala, Aila, and Laine]{edm}
Tero Karras, Miika Aittala, Timo Aila, and Samuli Laine.
\newblock Elucidating the design space of diffusion-based generative models.
\newblock In \emph{NeurIPS}, 2022.

\bibitem[Kasten et~al.(2021)Kasten, Ofri, Wang, and Dekel]{atlas}
Yoni Kasten, Dolev Ofri, Oliver Wang, and Tali Dekel.
\newblock Layered neural atlases for consistent video editing.
\newblock \emph{ACM Transactions on Graphics (TOG)}, 40\penalty0 (6):\penalty0 1--12, 2021.

\bibitem[Khachatryan et~al.(2023)Khachatryan, Movsisyan, Tadevosyan, Henschel, Wang, Navasardyan, and Shi]{text2videozero}
Levon Khachatryan, Andranik Movsisyan, Vahram Tadevosyan, Roberto Henschel, Zhangyang Wang, Shant Navasardyan, and Humphrey Shi.
\newblock Text2video-zero: Text-to-image diffusion models are zero-shot video generators.
\newblock \emph{arXiv preprint arXiv:2303.13439}, 2023.

\bibitem[Kingma and Ba(2014)]{adam}
Diederik~P Kingma and Jimmy Ba.
\newblock Adam: A method for stochastic optimization.
\newblock \emph{arXiv preprint arXiv:1412.6980}, 2014.

\bibitem[Kirillov et~al.(2023)Kirillov, Mintun, Ravi, Mao, Rolland, Gustafson, Xiao, Whitehead, Berg, Lo, et~al.]{sam}
Alexander Kirillov, Eric Mintun, Nikhila Ravi, Hanzi Mao, Chloe Rolland, Laura Gustafson, Tete Xiao, Spencer Whitehead, Alexander~C Berg, Wan-Yen Lo, et~al.
\newblock Segment anything.
\newblock \emph{arXiv preprint arXiv:2304.02643}, 2023.

\bibitem[Ling et~al.(2023)Ling, Chen, Zhang, Chen, and Jin]{freedrag}
Pengyang Ling, Lin Chen, Pan Zhang, Huaian Chen, and Yi Jin.
\newblock Freedrag: Point tracking is not you need for interactive point-based image editing.
\newblock \emph{arXiv preprint arXiv:2307.04684}, 2023.

\bibitem[Liu et~al.(2017)Liu, Yeh, Tang, Liu, and Agarwala]{smooth4}
Ziwei Liu, Raymond~A Yeh, Xiaoou Tang, Yiming Liu, and Aseem Agarwala.
\newblock Video frame synthesis using deep voxel flow.
\newblock In \emph{Proceedings of the IEEE international conference on computer vision}, pages 4463--4471, 2017.

\bibitem[Luo(2022)]{unified_perspective_diffusion}
Calvin Luo.
\newblock Understanding diffusion models: {A} unified perspective.
\newblock \emph{CoRR}, abs/2208.11970, 2022.

\bibitem[Midjourney(2023)]{Midjourney}
Midjourney.
\newblock Midjourney, 2023.

\bibitem[Mou et~al.(2023)Mou, Wang, Song, Shan, and Zhang]{dragondiffusion}
Chong Mou, Xintao Wang, Jiechong Song, Ying Shan, and Jian Zhang.
\newblock Dragondiffusion: Enabling drag-style manipulation on diffusion models.
\newblock \emph{arXiv preprint arXiv:2307.02421}, 2023.

\bibitem[M{\"u}ller et~al.(2022)M{\"u}ller, Evans, Schied, and Keller]{instantngp}
Thomas M{\"u}ller, Alex Evans, Christoph Schied, and Alexander Keller.
\newblock Instant neural graphics primitives with a multiresolution hash encoding.
\newblock \emph{ACM Transactions on Graphics (ToG)}, 41\penalty0 (4):\penalty0 1--15, 2022.

\bibitem[Nichol and Dhariwal(2021)]{ddpm2}
Alexander~Quinn Nichol and Prafulla Dhariwal.
\newblock Improved denoising diffusion probabilistic models.
\newblock In \emph{{ICML}}, pages 8162--8171. {PMLR}, 2021.

\bibitem[OpenAI(2023)]{Dalle-2}
OpenAI.
\newblock Dalle-2, 2023.

\bibitem[Pan et~al.(2023)Pan, Tewari, Leimk{\"u}hler, Liu, Meka, and Theobalt]{draggan}
Xingang Pan, Ayush Tewari, Thomas Leimk{\"u}hler, Lingjie Liu, Abhimitra Meka, and Christian Theobalt.
\newblock Drag your gan: Interactive point-based manipulation on the generative image manifold.
\newblock In \emph{ACM SIGGRAPH 2023 Conference Proceedings}, pages 1--11, 2023.

\bibitem[Perazzi et~al.(2016)Perazzi, Pont-Tuset, McWilliams, Van~Gool, Gross, and Sorkine-Hornung]{davis}
Federico Perazzi, Jordi Pont-Tuset, Brian McWilliams, Luc Van~Gool, Markus Gross, and Alexander Sorkine-Hornung.
\newblock A benchmark dataset and evaluation methodology for video object segmentation.
\newblock In \emph{Proceedings of the IEEE conference on computer vision and pattern recognition}, pages 724--732, 2016.

\bibitem[Qi et~al.(2023)Qi, Cun, Zhang, Lei, Wang, Shan, and Chen]{fatezero}
Chenyang Qi, Xiaodong Cun, Yong Zhang, Chenyang Lei, Xintao Wang, Ying Shan, and Qifeng Chen.
\newblock Fatezero: Fusing attentions for zero-shot text-based video editing.
\newblock \emph{arXiv preprint arXiv:2303.09535}, 2023.

\bibitem[Radford et~al.(2021)Radford, Kim, Hallacy, Ramesh, Goh, Agarwal, Sastry, Askell, Mishkin, Clark, Krueger, and Sutskever]{clip}
Alec Radford, Jong~Wook Kim, Chris Hallacy, Aditya Ramesh, Gabriel Goh, Sandhini Agarwal, Girish Sastry, Amanda Askell, Pamela Mishkin, Jack Clark, Gretchen Krueger, and Ilya Sutskever.
\newblock Learning transferable visual models from natural language supervision.
\newblock In \emph{{ICML}}, pages 8748--8763. {PMLR}, 2021.

\bibitem[Rombach et~al.(2022)Rombach, Blattmann, Lorenz, Esser, and Ommer]{stable_diffusion}
Robin Rombach, Andreas Blattmann, Dominik Lorenz, Patrick Esser, and Bj{\"{o}}rn Ommer.
\newblock High-resolution image synthesis with latent diffusion models.
\newblock In \emph{{CVPR}}, pages 10674--10685. {IEEE}, 2022.

\bibitem[Saharia et~al.(2022)Saharia, Chan, Saxena, Li, Whang, Denton, Ghasemipour, Lopes, Ayan, Salimans, Ho, Fleet, and Norouzi]{imagen}
Chitwan Saharia, William Chan, Saurabh Saxena, Lala Li, Jay Whang, Emily~L. Denton, Seyed Kamyar~Seyed Ghasemipour, Raphael~Gontijo Lopes, Burcu~Karagol Ayan, Tim Salimans, Jonathan Ho, David~J. Fleet, and Mohammad Norouzi.
\newblock Photorealistic text-to-image diffusion models with deep language understanding.
\newblock In \emph{NeurIPS}, 2022.

\bibitem[Shi et~al.(2023)Shi, Xue, Pan, Zhang, Tan, and Bai]{dragdiffusion}
Yujun Shi, Chuhui Xue, Jiachun Pan, Wenqing Zhang, Vincent~YF Tan, and Song Bai.
\newblock Dragdiffusion: Harnessing diffusion models for interactive point-based image editing.
\newblock \emph{arXiv preprint arXiv:2306.14435}, 2023.

\bibitem[Singer et~al.(2022)Singer, Polyak, Hayes, Yin, An, Zhang, Hu, Yang, Ashual, Gafni, et~al.]{make-a-video}
Uriel Singer, Adam Polyak, Thomas Hayes, Xi Yin, Jie An, Songyang Zhang, Qiyuan Hu, Harry Yang, Oron Ashual, Oran Gafni, et~al.
\newblock Make-a-video: Text-to-video generation without text-video data.
\newblock \emph{arXiv preprint arXiv:2209.14792}, 2022.

\bibitem[Song et~al.(2021)Song, Meng, and Ermon]{ddim}
Jiaming Song, Chenlin Meng, and Stefano Ermon.
\newblock Denoising diffusion implicit models.
\newblock In \emph{{ICLR}}. OpenReview.net, 2021.

\bibitem[Tang et~al.(2023)Tang, Jia, Wang, Phoo, and Hariharan]{dift}
Luming Tang, Menglin Jia, Qianqian Wang, Cheng~Perng Phoo, and Bharath Hariharan.
\newblock Emergent correspondence from image diffusion.
\newblock \emph{arXiv preprint arXiv:2306.03881}, 2023.

\bibitem[Teed and Deng(2020)]{raft}
Zachary Teed and Jia Deng.
\newblock Raft: Recurrent all-pairs field transforms for optical flow.
\newblock In \emph{Computer Vision--ECCV 2020: 16th European Conference, Glasgow, UK, August 23--28, 2020, Proceedings, Part II 16}, pages 402--419. Springer, 2020.

\bibitem[Wang et~al.(2023{\natexlab{a}})Wang, Yang, Tuo, He, Zhu, Fu, and Liu]{wang2023videofactory}
Wenjing Wang, Huan Yang, Zixi Tuo, Huiguo He, Junchen Zhu, Jianlong Fu, and Jiaying Liu.
\newblock Videofactory: Swap attention in spatiotemporal diffusions for text-to-video generation, 2023{\natexlab{a}}.

\bibitem[Wang et~al.(2023{\natexlab{b}})Wang, Yuan, Zhang, Chen, Wang, Zhang, Shen, Zhao, and Zhou]{videocomposer}
Xiang Wang, Hangjie Yuan, Shiwei Zhang, Dayou Chen, Jiuniu Wang, Yingya Zhang, Yujun Shen, Deli Zhao, and Jingren Zhou.
\newblock Videocomposer: Compositional video synthesis with motion controllability.
\newblock \emph{arXiv preprint arXiv:2306.02018}, 2023{\natexlab{b}}.

\bibitem[Wu et~al.(2023{\natexlab{a}})Wu, Ge, Wang, Lei, Gu, Shi, Hsu, Shan, Qie, and Shou]{tuneavideo}
Jay~Zhangjie Wu, Yixiao Ge, Xintao Wang, Stan~Weixian Lei, Yuchao Gu, Yufei Shi, Wynne Hsu, Ying Shan, Xiaohu Qie, and Mike~Zheng Shou.
\newblock Tune-a-video: One-shot tuning of image diffusion models for text-to-video generation.
\newblock In \emph{Proceedings of the IEEE/CVF International Conference on Computer Vision}, pages 7623--7633, 2023{\natexlab{a}}.

\bibitem[Wu et~al.(2023{\natexlab{b}})Wu, Li, Gao, Dong, Bai, Singh, Xiang, Li, Huang, Sun, et~al.]{loveu_edit}
Jay~Zhangjie Wu, Xiuyu Li, Difei Gao, Zhen Dong, Jinbin Bai, Aishani Singh, Xiaoyu Xiang, Youzeng Li, Zuwei Huang, Yuanxi Sun, et~al.
\newblock Cvpr 2023 text guided video editing competition.
\newblock \emph{arXiv preprint arXiv:2310.16003}, 2023{\natexlab{b}}.

\bibitem[Zhang et~al.(2023)Zhang, Wei, Jiang, Zhang, Zuo, and Tian]{controlvideo}
Yabo Zhang, Yuxiang Wei, Dongsheng Jiang, Xiaopeng Zhang, Wangmeng Zuo, and Qi Tian.
\newblock Controlvideo: Training-free controllable text-to-video generation.
\newblock \emph{arXiv preprint arXiv:2305.13077}, 2023.

\bibitem[Zhou et~al.(2022)Zhou, Wang, Yan, Lv, Zhu, and Feng]{zhou2022magicvideo}
Daquan Zhou, Weimin Wang, Hanshu Yan, Weiwei Lv, Yizhe Zhu, and Jiashi Feng.
\newblock Magicvideo: Efficient video generation with latent diffusion models.
\newblock \emph{arXiv preprint arXiv:2211.11018}, 2022.

\end{thebibliography}
}
\clearpage
\setcounter{page}{1}
\appendix

\section{Quantitative Analysis}

We report the quantitative results in~\cref{tab:eva}. Lower values indicate better temporal smoothness.
We design an evaluation metric to evaluate the temporal smoothness of videos. 
The smoothness metric is based on the hypothesis that any pixel in a video moves linearly within a short time span, inspired by previous work~\cite{smooth1,smooth2,smooth3,smooth4}. 
Specifically, we first compute the optical flow~\cite{raft} among three neighboring frames (indexed as $i-1$, $i$, $i+1$) to represent the motion.
Then, for each pixel in the $i$-th frame, we calculate its distance to the line segment between the corresponding pixels on the $(i-1)$-th frame and the $(i+1)$-th frame. This distance represents the smoothness of the pixel motion, and it can be calculated in any videos, including the input videos and the edited videos.
For each pixel, We filter out the pixels whose distances on the edited video smaller than their distances on the input video.
Finally, We calculate the average of the remaining distances over all pixels and frames. This value is taken as a measurement of the temporal smoothness of edited videos. Lower values indicate better temporal smoothness.

\begin{table} 
    \centering
    \small
    \begin{tabular}{c|c}
    \toprule
    Method  & Smoothness $\downarrow$ \\ 
    \midrule
      Baseline & 1.42 \\
      Ours  & \textbf{1.11}   \\
    \bottomrule
    \end{tabular}
    \vspace{-.5em}
    \caption{Evaluation on temporal smoothness. Lower values indicate better temporal smoothness. }
    \vspace{-.5em}
    \label{tab:eva}
\end{table}

\section{Additional ablation studies}

\begin{figure}[t]
\centering
\includegraphics[width=0.99\linewidth]{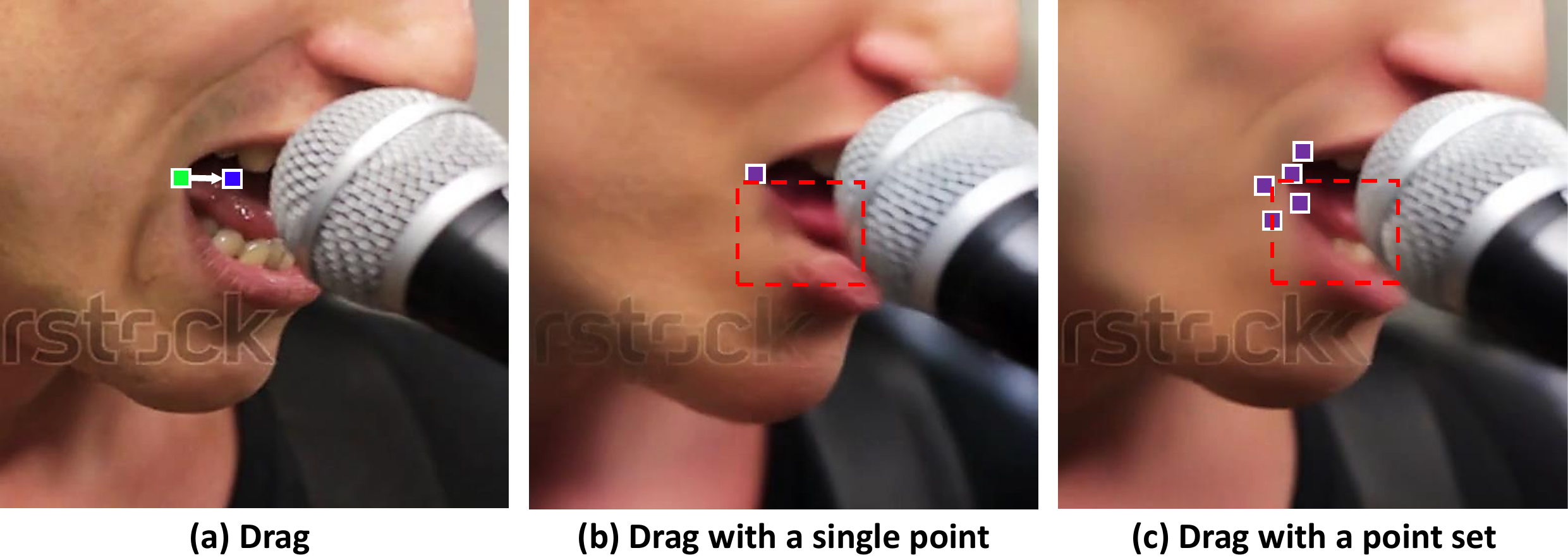}
\vspace{-.5em}
\caption{
Study of the point set for object deformation. The single handle point (the squares marked in purple) is located only at the corner of the lip, leaving part of lip generated to other content. The updated points from the handle point set can be located within the red dashed box, thus potentially preserving part of the lip.
} 
\vspace{-.5em}
\label{fig:abla_point_set_deform}
\end{figure}

\noindent
\textbf{Study on the point set for object deformation}. 
When a foreground object deforms across frames, the single handle point is unlikely to cover the same contents on all frames.
As shown in~\cref{fig:abla_point_set_deform}, when the singer opens the mouth, the points from the set still cover the lip, thus maintaining the integrity of the entire lip.
However, the single point is only located on the corner of mouth, so the other areas of the mouth may change arbitrarily, causing the lip to disappear.

\section{More results}
\begin{figure*}
    \centering
    \vspace{-.2em}\includegraphics[width=0.95\linewidth]{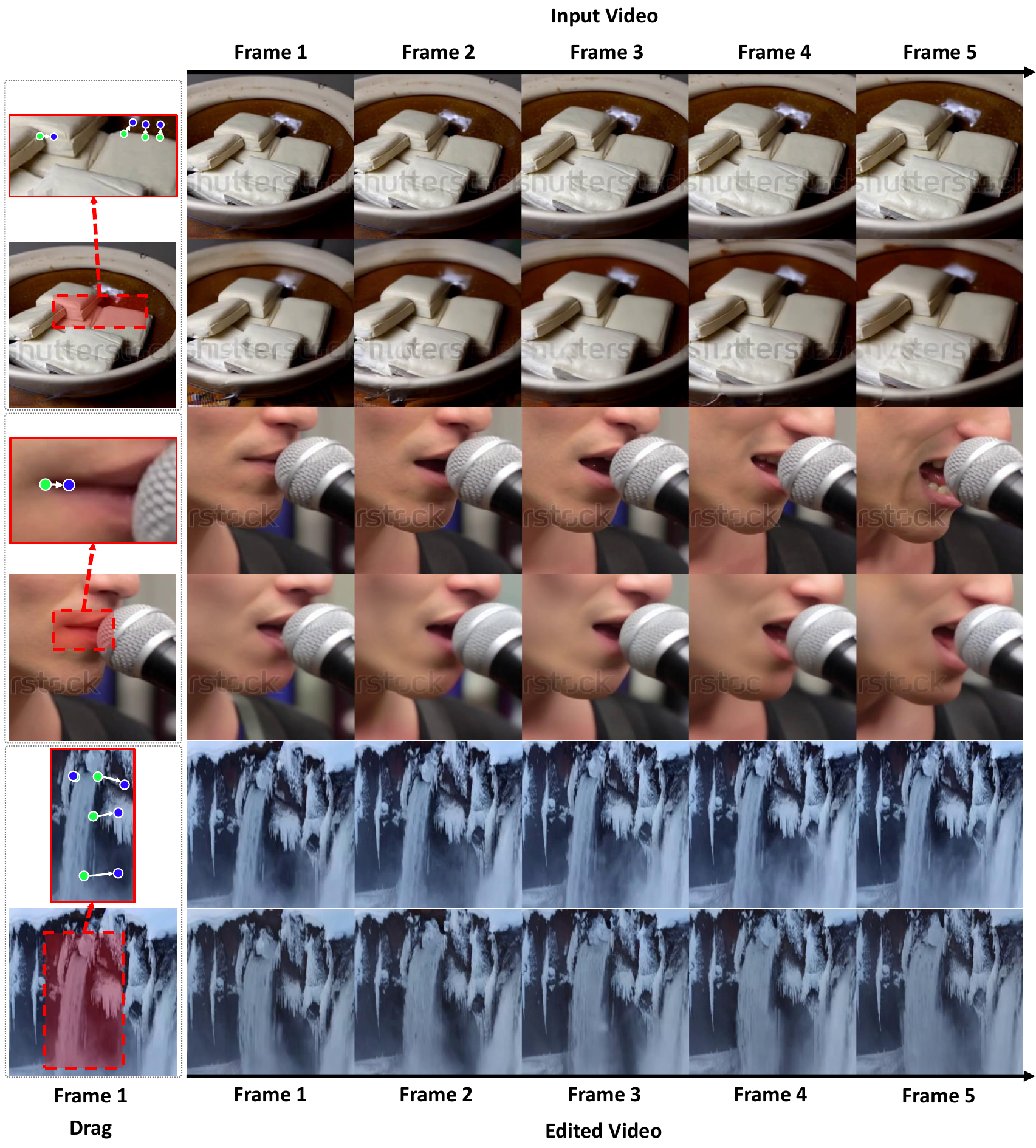}
    \vspace{-.7em}
    \caption{
    {\textbf{Results}}. The video samples were edited by our method. The green and blue points denote the handle and target points, respectively. The red point in the last case denotes the mask point.
    }
    \vspace{-1em}
    \label{fig:sota2}
\end{figure*}

\begin{figure*}
    \centering
    \vspace{-.2em}\includegraphics[width=0.95\linewidth]{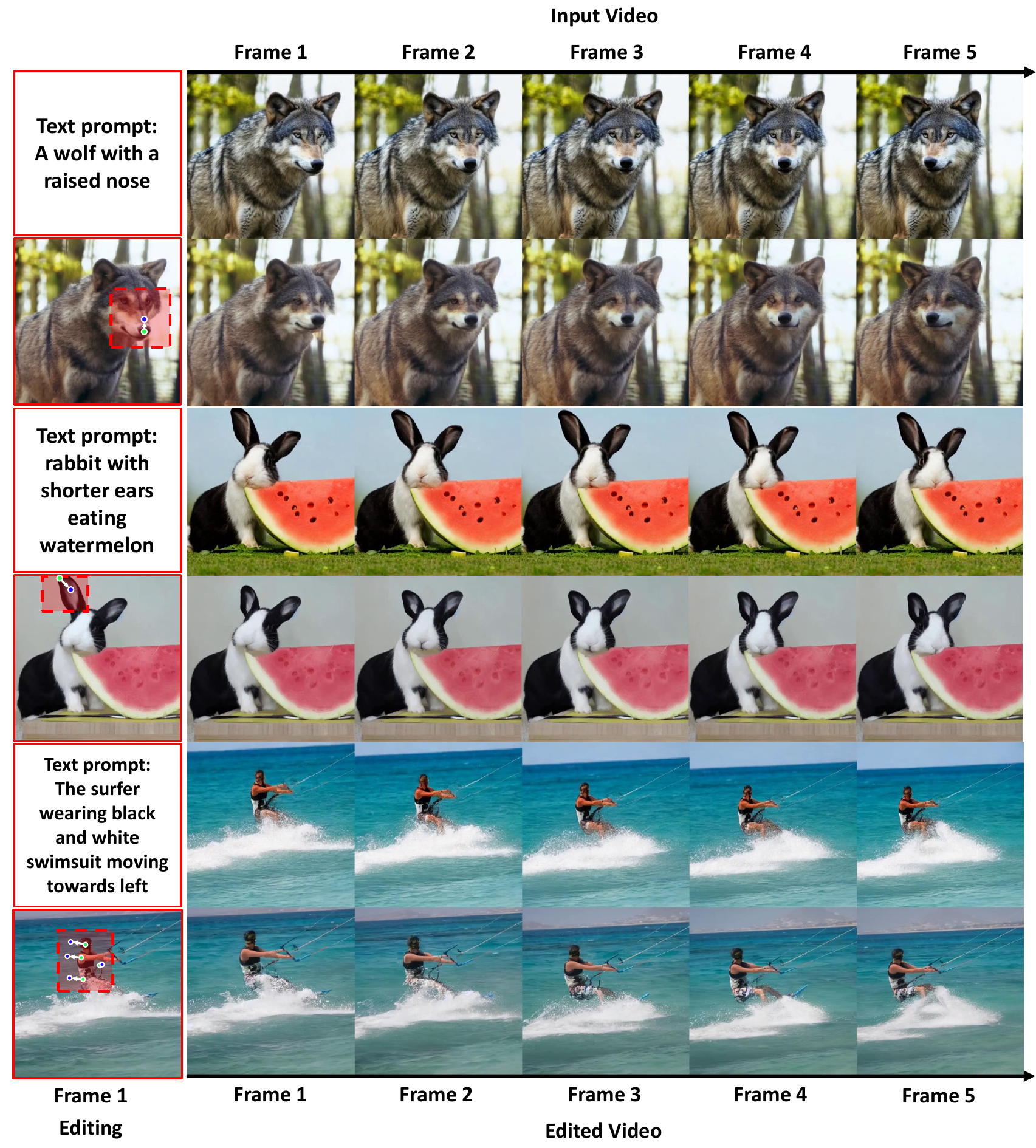}
    \vspace{-.7em}
    \caption{
    {\textbf{Results}}. The video samples were edited by our method. The green and blue points denote the handle and target points, respectively. The red point in the last case denotes the mask point. The first column is the way of editing. TokenFlow~\cite{tokenflow} uses the text prompt and our method uses the drag signal.
    }
    \vspace{-1em}
    \label{fig:sota_tokenflow}
\end{figure*}

We include more video samples in~\cref{fig:sota2}. 
As shown in~\cref{fig:sota_tokenflow}, we compare our method with a text-driven video editing method, TokenFlow~\cite{tokenflow}, to demonstrate the necessity of our framework. 
Although TokenFlow can achieve highly consistent results, it cannot deform the instances given the expected text prompt. Our framework with point-based guidance can achieve the desired results.

\section{Limitations}
Drag-based video editing is a challenging task, and there is still a lot of room for improvement of our method.
\begin{itemize}
    \item The 2D point propagations do not work perfectly in every scene. First, some points clicked on the key frames could be occluded in subsequent frames. Second, only 2D points cannot represent all the deformations on instances. These 2D points lack the depth information, so the manipulated component on instances is ambiguous. One possible solution is lifting the 2D points into 3D space, but in this case, a well pre-trained 3D diffusion model is required. 
    \item According to our observation, our framework is sensitive to the user input. The masks may not coordinate with the handle points because the change of handle points could change the overall structure of frames. Unfortunately, the current diffusion models have difficulty in self-calibrating all the generated areas to be temporally consistent like the original video~(not smooth and inconsistent with the laws of physical motion). Thus, the input masks are still necessary for the current video-level point-based manipulation.
\end{itemize} 

\end{document}